\documentclass{article} 
\usepackage{iclr2024_conference,times}

\usepackage{amsmath,amsfonts,bm}

\def\eqref#1{equation~\ref{#1}}

\def\1{\bm{1}}

\DeclareMathAlphabet{\mathsfit}{\encodingdefault}{\sfdefault}{m}{sl}
\SetMathAlphabet{\mathsfit}{bold}{\encodingdefault}{\sfdefault}{bx}{n}

\usepackage{hyperref}
\usepackage{url}
\usepackage{graphicx}

\usepackage{subcaption}
\usepackage{booktabs}

\title{3D-GPT: Procedural 3D Modeling with Large Language Models}

\author{Chunyi Sun$^{1 *}$, Junlin Han$^{2,3 *}$, Weijian Deng$^{1}$, Xinlong Wang$^{4}$, Zishan Qin$^{1}$, Stephen Gould$^{1}$  \\
$^{1}$Australian National University, $^{2}$University of Oxford, $^{3}$Cybever, $^{4}$Beijing Academy of Artificial Intelligence \\
\hspace{2cm}Project page: \textcolor{black}{\href{https://chuny1.github.io/3DGPT/3dgpt.html}{https://chuny1.github.io/3DGPT/3dgpt.html}}\\
\hspace{5cm}* Equal contribution \\
}

\begin{document}

\maketitle

\begin{abstract}

In the pursuit of efficient automated content creation, procedural generation, leveraging modifiable parameters and rule-based systems, emerges as a promising approach. Nonetheless, it could be a demanding endeavor, given its intricate nature necessitating a deep understanding of rules, algorithms, and parameters. To reduce workload, we introduce 3D-GPT, a framework utilizing large language models~(LLMs) for instruction-driven 3D modeling. 3D-GPT positions LLMs as proficient problem solvers, dissecting the procedural 3D modeling tasks into accessible segments and appointing the apt agent for each task. 3D-GPT integrates three core agents: the task dispatch agent, the conceptualization agent, and the modeling agent. They collaboratively achieve two objectives. First, it enhances concise initial scene descriptions, evolving them into detailed forms while dynamically adapting the text based on subsequent instructions. Second, it integrates procedural generation, extracting parameter values from enriched text to effortlessly interface with 3D software for asset creation. Our empirical investigations confirm that 3D-GPT not only interprets and executes instructions, delivering reliable results but also collaborates effectively with human designers. Furthermore, it seamlessly integrates with Blender, unlocking expanded manipulation possibilities. 
Our work highlights the potential of LLMs in 3D modeling, offering a basic framework for future advancements in scene generation and animation.

\end{abstract}


\section{Introduction} \label{sec:intro}

In the metaverse era, 3D content creation serves as a catalyst for transformative progress, redefining multimedia experiences in domains like gaming, virtual reality, and cinema with intricately crafted models.
Yet, designers often grapple with a time-intensive 3D modeling process, starting from basic shapes (\textit{e.g.}, cubes, spheres, or cylinders) and employing software like Blender for meticulous shaping, detailing, and texturing. This demanding workflow concludes with rendering and post-processing to deliver the polished final model.
While procedural generation holds promise with its efficiency in automating content creation through adjustable parameters and rule-based systems~\citep{procthor,greff2022kubric,he2021semi,jiang2018configurable,infinigen}, it demands a comprehensive grasp of generation rules, algorithmic frameworks, and individual parameters. 
Furthermore, aligning these processes with the creative visions of clients, through effective communication, adds another layer of complexity.
This underscores the importance of simplifying the traditional 3D modeling workflow to empower creators in the metaverse era.

LLMs have showcased exceptional language understanding capabilities, including planning and tool utilization~\citep{imani2023mathprompter,zhang2023proagent,gong2023mindagent,zeng2022socratic}. 

Furthermore, LLMs demonstrate outstanding proficiency in characterizing object attributes, such as structure and texture~\citep{menon2022visual,pratt2022does,fan2023improving}, enabling them to enhance details from rough descriptions. Additionally, they excel at parsing concise textual information and comprehending intricate code functions, while seamlessly facilitating efficient interactions with users.
Driven by these extraordinary capabilities, we embark on exploring their innovative applications in procedural 3D modeling. Our primary objective is to harness the power of LLMs to exert control over 3D creation software in accordance with the requirements of clients.

In pursuit of this vision, we introduce 3D-GPT, a framework aimed at facilitating instruction-driven 3D content synthesis. 3D-GPT enables LLMs to function as problem-solving agents, breaking down the 3D modeling task into smaller, manageable components, and determining when, where, and how to accomplish each segment.
3DGPT comprises three key agents: conceptualization agent, 3D modeling agent and task dispatch agent. The first two agents collaborate harmoniously to fulfill the roles of 3D conceptualization and 3D modeling by manipulating the 3D generation functions. Subsequently, the third agent manages the system by taking the initial text input, handling sub-sequence instructions, and facilitating effective cooperation between the two aforementioned agents.

By doing so, they work toward two key objectives. First, it enhances initial scene descriptions, guiding them towards more detailed and contextually relevant forms while adapting the textual input based on subsequent instructions. Second, instead of directly crafting every element of 3D content, we employ procedural generation, making use of adaptable parameters and rule-based systems to interface with 3D software. Our 3D-GPT is equipped with the capability to understand procedural generation functions and extract corresponding parameter values from the enriched text. 

3D-GPT offers controllable and precise 3D generation guided by users' textual descriptions. It reduces the workload of manually defining each controllable parameter in procedural generation, particularly within complex scenes that encompass diverse aspects.
Moreover, 3D-GPT enhances collaboration with users, making the creative process more efficient and user-centric.
Furthermore, 3D-GPT seamlessly interfaces with Blender, granting users diverse manipulation capabilities: object transformations, material adjustments, primitive additions, object animations, mesh editing, and physical motion simulations.
Based on our experiments, we posit that LLMs exhibit the potential to handle more intricate visual inputs. Our contributions are summarized as follows:
\begin{itemize}
  \vspace{-5pt}
  \itemsep -3pt\partopsep -7pt
     \item Introducing 3D-GPT, a training-for-free framework designed for 3D scene generation. Our approach leverages the innate multimodal reasoning capabilities of LLMs, streamlining the efficiency of end-users engaged in procedural 3D modeling.
    \item Exploration of an alternative path in text-to-3D generation, wherein our 3D-GPT generates Python codes to control 3D software, potentially offering increased flexibility for real-world applications.
    \item Empirical experiments demonstrate the substantial potential of LLMs in terms of their reasoning, planning, and tool-using capabilities in 3D content generation.
\end{itemize}

\section{Related Work}
\label{sec:rw}

\subsection{Text-to-3D generation}
With the recent advance in text-to-image generation modeling, there has been a growing interest in text-to-3D generation~\citep{sanghi2022clip,poole2022dreamfusion,magic3d,xu2023dream3d,latentnerf,prolificdreamer,xu2023dream3d,mohammad2022clip,jain2022zero}. The common paradigm of them is to perform per-shape optimization with differentiable rendering and the guidance of the CLIP model~\citep{radford2021learning} or 2D diffusion models~\citep{rombach2022high}. For example, 
DreamFields~\citep{jain2022zero} and CLIP-Mesh~\citep{mohammad2022clip} explore zero-shot 3D content creation using only CLIP guidance.
Dreamfusion~\citep{poole2022dreamfusion} optimizes NeRF~\cite{nerf} with the guidance of a text-to-image diffusion model, achieving remarkable text-to-3D synthesis results.
To address optimization speed and visual quality challenges, Magic3D~\citep{magic3d} uses low-resolution diffusion priors and a sparse 3D hash grid for speed, alongside an efficient differentiable render for textured 3D mesh model optimization.
To enhance the fidelity of generated 3D models, innovative approaches refine the 3D geometry. For instance, Dream3D~\citep{xu2023dream3d} directly initializes NeRF using a generated Signed Distance Function (SDF) for better geometry control. Latent-NeRF~\citep{latentnerf} incorporates a user-provided mesh for direct occupancy loss during geometry optimization in NeRF. 
Subject-driven text-to-3D generation is gaining traction for personalized synthesis~\citep{raj2023dreambooth3d,liu2023one, melas2023realfusion}. It creates subject-specific 3D assets based on input images and text prompts.
To achieve this, DreamBooth3D~\citep{raj2023dreambooth3d} proposes a 3-stage optimization strategy to jointly leverage the 3D consistency of NeRF together with the personalizing capability of the text-to-image diffusion model. One-2-3-45~\citep{liu2023one} uses a view-conditioned 2D diffusion model (Zero123) to generate multi-view images for learning SDF-based generalizable neural surface reconstruction.
Unlike the above approaches, our objective is not to generate conventional neural representations as the final 3D output. Instead, we utilize LLMs to generate Python code that controls Blender's 3D modeling based on the provided instructions.

\subsection{Large language models}
Large language models (LLMs) are a promising approach to capture and represent the compressed knowledge and experiences of humans, projecting them into language space~\citep{devlin2018bert, 2020t5, openai2023gpt4,chowdhery2022palm,bubeck2023sparks}. LLMs have consistently showcased remarkable performance extending beyond canonical language processing domains. They exhibit the capability to address intricate tasks that were once considered the exclusive domain of specialized algorithms or human experts. These tasks encompass areas such as mathematical reasoning~\citep{imani2023mathprompter,wei2022chain}, medicine~\citep{jeblick2022chatgpt, yang2023evaluations}, and planning~\citep{zhang2023proagent,gong2023mindagent,huang2023voxposer,huang2022language}
For instance, \citet{huang2022language} leverage the LLMs' internet-scale domain knowledge and emergent zero-shot planning abilities to perform complex task planning and reasoning.
\citet{gong2023mindagent} explore LLMs in multi-agent coordination in scenarios encompassing multiple task objectives.
\citep{zeng2022socratic} introduce a modular framework that leverages structured dialogue via prompting between multiple large pretrained models to make joint predictions for new multimodal tasks, without requiring finetuning.
Moreover, specialized LLMs for particular applications have been explored such as Codex~\citep{chen2021evaluating} for Python code generation, Galactica~\citep{taylor2022galactica} for scientific knowledge, and LaMDA~\citep{thoppilan2022lamda} for dialogue applications.
This work explores the innovative application of LLMs in 3D modeling,  employing them to control 3D procedural generation.
\section{3D-GPT}

\subsection{Task Formulation}

The overall objective is the generation of 3D content based on a sequence of natural language instructions, denoted as ${\cal L}=[L_i]$. The initial instruction, designated as $L_0$, serves as a comprehensive description of the 3D scene, such as ``\textit{A misty spring morning, where dew-kissed flowers dot a lush meadow surrounded by budding trees}". Subsequent instructions are employed to modify the existing scene, as exemplified by instructions like ``\textit{Transform the white flowers into yellow flower}" or ``\textit{translate the scene into a winter setting}".

To accomplish this objective, we introduce a framework named 3D-GPT, which empowers LLMs to act as problem-solving agents. We point out that employing LLMs to directly create every element of 3D content poses significant challenges. LLMs lack specific pre-training data for proficient 3D modeling and, as a result, may struggle to accurately determine which elements to modify and how to modify them based on given instructions.

To avoid this challenge, we employ procedural generation to control the 3D content creation. It makes use of adaptable parameters and rule-based systems to interface with 3D software (\textit{e.g.}, Blender) so as to efficiently  conduct 3D modeling~\citep{procthor,greff2022kubric,he2021semi,jiang2018configurable,infinigen}.

Our approach conceptualizes the 3D procedural generation engine as a set of functions, denoted as ${\cal F} = \{F_j\}$, where each function $F_j$ takes parameters $P_j$ as input.

 Within the 3D-GPT framework, for each instruction $L_i$, we frame the modeling task as the selection of a subset $\hat{\cal F} \subseteq {\cal F}$, combined with the inference of corresponding parameters $P_j$ for each function~$F_j$ in this subset. The ultimate aim is to ensure that the $\hat{\cal F}$ collaboratively generates a 3D scene that aligns with the descriptions provided in~${\cal L}$. By adeptly addressing both function selection and parameter inference for every sub-instruction $L_i$, 3D-GPT generates a Python script file with the capacity to manipulate Blender's 3D modeling environment, thereby proficiently meeting the user's specified requirements outlined in the instruction sequence~${\cal L}$.

\subsection{Modeling Tool Preparation}
\label{modeling_tool_preparation}
In our framework, we utilize Infinigen~\cite{infinigen}, a Python-Blender-based procedural generator equipped with a rich library of generation functions. To empower LLMs with the ability to proficiently leverage Infinigen, we provide crucial prompts for each function $F_j$. These prompts encompass function documentation $D_j$, easily understandable code $C_j$, required information $I_j$, and a usage example $E_j$: 

\begin{itemize}
  \vspace{-5pt}
  \itemsep -3pt\partopsep -7pt

\item $D_j$: it entails a comprehensive explanation of the input parameter $P_j$, coupled with a clear elucidation of the function's purpose and functionality.
\item $C_j$: we present meticulously restructured and highly readable function code, ensuring that it is accessible and comprehensible for LLMs.
\item $I_j$: this component outlines the specific information required to infer the function parameters, thereby assisting LLMs in understanding the context and prerequisites of each function. For example, in the case of a flower generation function, $I_j$ indicates the required visual properties for modeling, such as flower color, flower petal appearance (\textit{e.g.}, size, curve, and length), and flower center appearance.
\item $E_j$: we provide illustrative examples to demonstrate how to infer the parameter $P_j$ from the accompanying text descriptions and subsequently invoke the function. Continuing with the example of a flower generation function, $E_j$ includes a practical demonstration of how to infer the parameters and call the function based on input text like ``a sunflower."
\end{itemize}
By providing LLMs with these resources, we enable them to leverage their core competencies in planning, reasoning, and tool utilization. As a result, LLMs can effectively harness Infinigen for 3D generation based on language instructions in a seamless and efficient manner.

\subsection{Multi-agents for 3D Reasoning, Planing and Tool Using}
\begin{figure*}[t]
    \centering
    \includegraphics[width=1\linewidth]{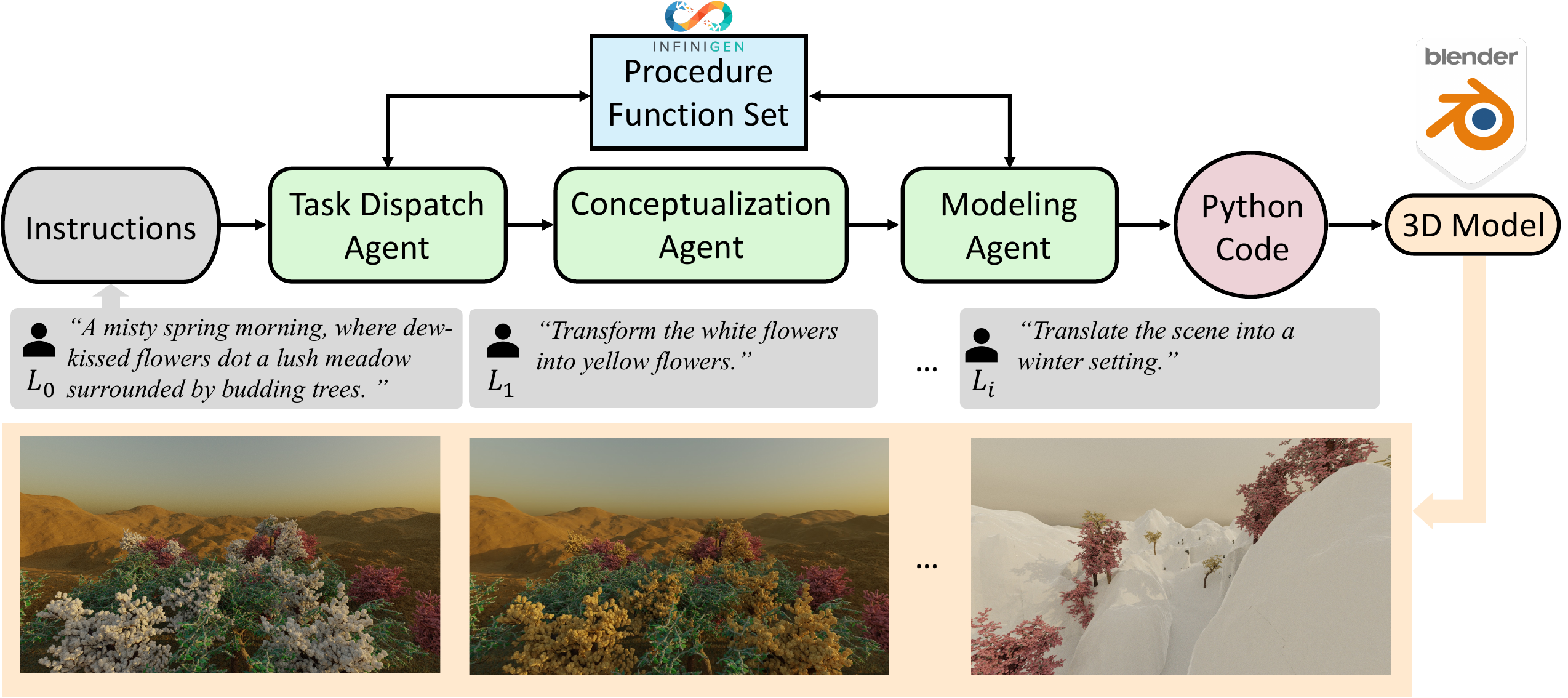}
    \caption{\textbf{3D-GPT Overview.} 3D-GPT employs LLMs as a multi-agent system with three collaborative agents for procedural 3D generation. These agents consult documents from the procedural generator, infer function parameters, and produce Python code. The generated code script interfaces with Blender's API for 3D content creation and rendering.}
    \vspace{-1cm}
    \label{fig:method_overview}
\end{figure*}

Upon tool preparation, 3D-GPT employs a multi-agent system to tackle the procedural 3D modeling task. This system comprises three integral agents: the task dispatch agent, the conceptualization agent, and the modeling agent, illustrated in Figure~\ref{fig:method_overview}. Together, they deconstruct the procedural 3D modeling task into manageable segments, with each agent specializing in distinct aspects: 3D reasoning, planning, and tool utilization.
The task dispatch agent plays a pivotal role in the planning process. It leverages user instructions to query function documents and subsequently selects the requisite functions for execution.
Once functions are selected, the conceptualization agent engages in reasoning to enrich the user-provided text description.
Building upon this, the modeling agent deduces the parameters for each selected function and generates Python code scripts to invoke Blender's API, facilitating the creation of the corresponding 3D content.
Additionally, images can be rendered using Blender rendering capability.

\textbf{Task Dispatch Agent for Planing.} The Task Dispatch Agent, armed with comprehensive information of all available functions ${\cal F}$ within the procedural generation, efficiently identifies the requisite functions for each instructional input. For instance, when presented with an instruction such as ``\textit{translate the scene into a winter setting}", it pinpoints functions like 
$\text{add\_snow\_layer}()$ and $\text{update\_trees}()$.
This pivotal role played by the task dispatch agent is instrumental in facilitating efficient task coordination between the conceptualization and modeling agents. Without it, the conceptualization and the modeling agents have to analyze all provided functions $\cal{F}$ for each given instruction.
This not only increases the workload for these agents but also extends processing time and can potentially lead to undesired modifications.

The communication flow between the LLM system, the user, and the task dispatch agent is outlined as follows:

\noindent\fbox{%
    \parbox{0.98\textwidth}{%
\textit{
\textbf{--- LLM System}: You are a proficient planner for selecting suitable functions based on user instructions. You are provided with the following functions: \small{$<(F_j ^\textit{\text{name}}, F_j ^\textit{\text{usage}})>$}. Below are a few examples of how to choose functions based on user instructions: \small{$<E_j^{\text{task\_dispatch}}>$}.}

\textit{\textbf{--- User}: My instruction is: \small{$<L_i>$}.}

\textit{\textbf{--- Task Dispatch Agent}: Given the instruction \small{$<L_i>$}, we determine the sublist of functions \small{$\hat{\cal F}$} that need to be used for 3D modeling.}
}}

In this context, $<(F_j ^\textit{\text{name}}, F_j ^\textit{\text{usage}})>$ represents a list of function names and concise function usage descriptions for all available functions and examples {$<E^{\textit{\text{task\_dispatch}}}>$} provide guided examples for prompt-based instructions.

\textbf{Engaging the Conceptualization Agent for Reasoning.}  The description may not explicitly provide the detailed appearance descriptions needed for modeling. For instance, consider the description: ``\textit{A misty spring morning, where dew-kissed flowers dot a lush meadow surrounded by budding trees}". When employing a tree modeling function that necessitates parameters such as tree branch length, tree size, and leaf type, it becomes apparent that these specific details are not directly stated in the given text.
When instructing the modeling agent to infer parameters directly, it tends to provide simplistic solutions, like using default or reasonable values from the parameter document or copying values from prompting examples. This reduces diversity in generation and complicates parameter inference.

To alleviate this issue, we introduce the conceptualization agent which collaborates with the task dispatch agent to augment the user-provided text description ($L_i$).
After the task dispatch agent selects the required functions, we send the user input text and the corresponding function-specific information to the conceptualization agent and request augmented text. For each function $F_j$, it enriches $L_i$ into detailed appearance descriptions $L^j_i$. 
The communication between the system and the Conceptualization Agent for instruction $<L_i>$ and function $<F_j>$ is as follows:

\noindent\fbox{%
    \parbox{0.98\textwidth}{%
\textit{\textbf{--- LLM System}: You are a skilled writer, especially when it comes to describing the appearance of objects and large scenes. Given a description $<L_i>$, provide detailed descriptions for the following information $<I_j>$. For terms not mentioned in the description, use your imagination to ensure they fit the text description.}

\textit{\textbf{--- Conceptualization Agent}: Given the \small{$<L_i>$} and requested information \small{$<I_j>$}, the extended description is: \small{$<\widehat{L_i^j}>$}.}
}}

\textbf{Modeling Agent for Tool Using.} After conceptualization, the 3D modeling processing is targeted to convert the detailed human language to machine-understandable language. 
 
In our framework, our modeling agent manipulates the functions of procedural modeling in the library to create a realistic 3D model. 
For each function $F_j$ and user instruction $L_i$, the task dispatch agent receive augmented context $\widehat{L_i}^j$ from the conceptualization agent. For each function $F_j$, we have the code $C_j$, function documentation $D_j$, and one usage example $E_j$. 
The modeling agent utilizes this information to select the appropriate functions and deduce the corresponding parameters. Subsequently, the modeling agent generates Python code that accurately calls the selected function (\textit{e.g.}, call it in the loop, not call) and correctly passes the inferred parameters with the appropriate data types to the function.

The two-turn utterances are based on the following pattern:

\noindent\fbox{%
    \parbox{0.98\textwidth}{%
        \textit{\textbf{--- LLM System}: You are a good 3D designer who can convert long text descriptions into parameters, and is good at understanding Python functions to manipulate 3D content. Given the text description~\small{$<\widehat{L_{i}^f}>$}, we have the following function codes \small{$<C_j>$} and the document for function~\small{$<D_j>$}. Below is an example bout how to make function calls to model the scene to fit the description: \small{$<E_j^{\text{\textit{modeling}}}>$}. Understand the function, and model the 3D scene that fits the text description by making a function call.}
       
        \textit{
        \textbf{--- Modeling Agent}: Given the description \small{$<\widehat{L_i^j}>$}, we use the following functions: ..., and their respective parameter values ... are adopted.
        }
    }%
}

\paragraph{Blender Rendering.} The Modeling agent ultimately supplies the Python function calls with inferred parameters, which are employed for Blender node control and rendering, resulting in the production of the final 3D mesh and RGB results.
\section{Experiments}

\begin{figure*}[t]
    \centering
    \includegraphics[width=1\linewidth]{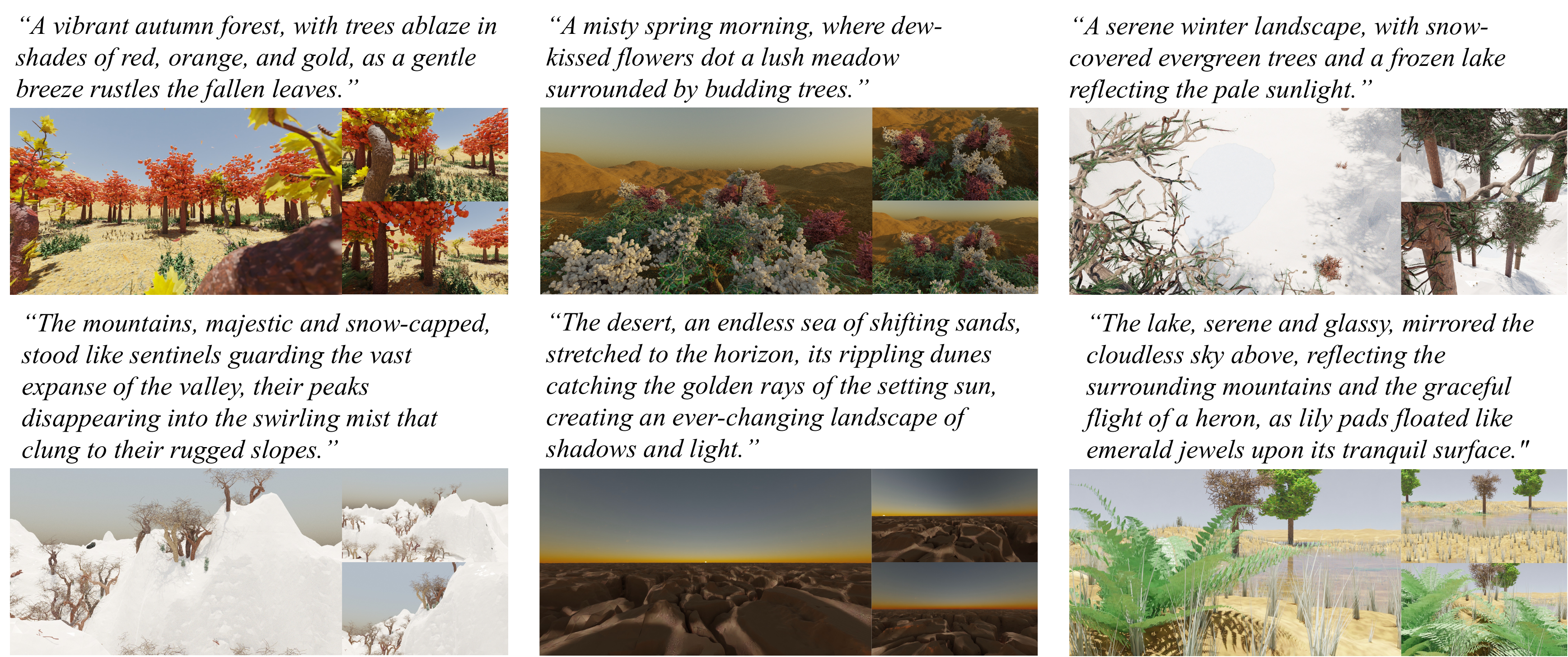}
    \caption{\textbf{Visual Examples of Instruction-Based 3D Scene Generation.} 3D-GPT can construct large 3D scenes that align with the provided initial instruction. We demonstrate that the rendered images contain various visual factors in line with the given instructions.}
    \label{fig:scene_generation}
\end{figure*}

Our experimentation begins by showcasing the proficiency of 3D-GPT in consistently generating results that align with user instructions, encompassing scenarios involving both large scenes and individual objects. Subsequently, we delve into specific examples to illustrate how our agents effectively comprehend tool functionalities, access necessary knowledge, and employ it for precise control.
To deepen our understanding, we conduct an ablation study to systematically examine the contributions of each agent within our multi-agent system.

\subsection{3D Modeling}

\paragraph{Large Scene Generation.} We investigate the capability of 3DGPT to control modeling tools based on scene descriptions \textit{without any training}. To conduct this experiment, we generated $100$ scene descriptions using ChatGPT with the following prompt: ``\textit{You are a good writer, provide $10$ different natural scene descriptions for me}". We collected responses to this prompt $10$ times to form our dataset.
In Figure~\ref{fig:scene_generation}, we present the multi-view rendering results of 3D-GPT. These results indicate that our approach is capable of generating large 3D scenes that generally align well with the provided text descriptions, showcasing a noticeable degree of diversity.
Notably, all 3D outcomes are directly rendered using Blender, ensuring that all meshes are authentic, thereby enabling our method to achieve absolute 3D consistency and produce real ray-tracing rendering results.

\paragraph{Fine-detail Control for Single Class.} 

\begin{figure*}[t]
    \vspace{-0.5cm}
    \centering
    \includegraphics[width=1\linewidth]{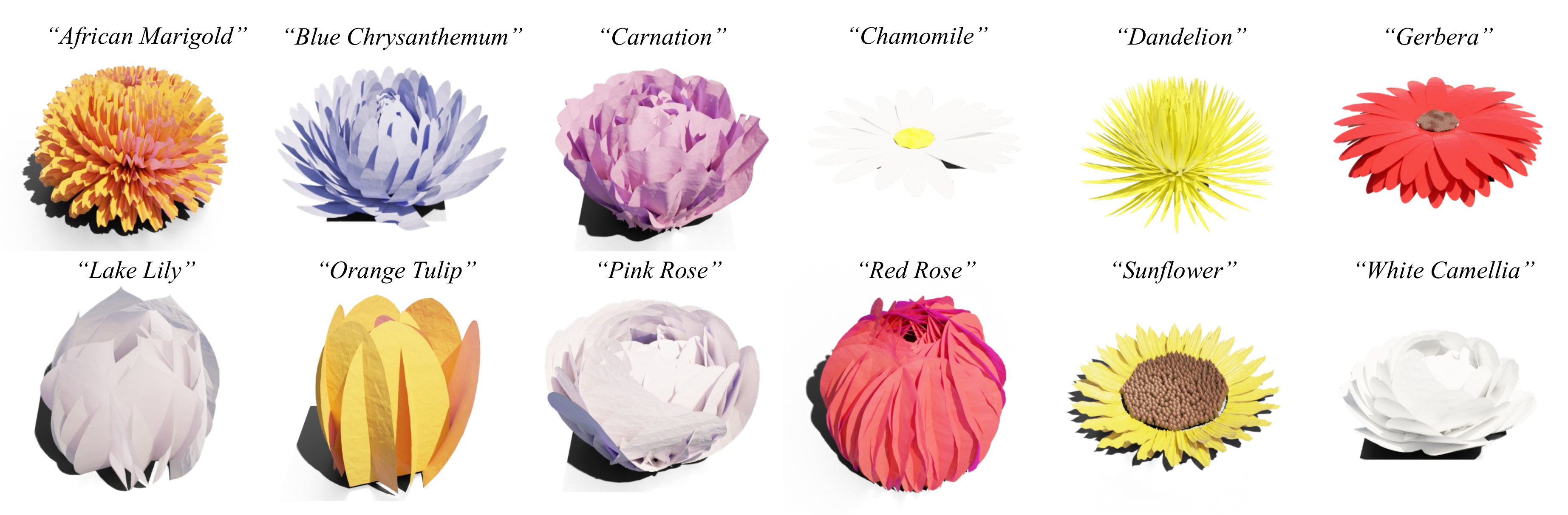}
    \caption{\textbf{Single Class Control Result.} Our method effectively acquires the necessary knowledge for modeling, enabling precise object control in terms of shape, curve, and key appearance capture. The generated results closely align with the given text. }
    \label{fig:single_class_result}
\end{figure*}

Apart from generating large scenes from concise descriptions, we assess the capabilities of 3D-GPT for modeling objects. We evaluate crucial factors such as curve modeling, shape control, and an in-depth understanding of object appearances. To this end, we report the results of fine-grained object control. This includes nuanced aspects such as object curves, key appearance features, and color, all derived from input text descriptions. We employ random prompts to instruct GPT for various real-world flower types. As depicted in Figure~\ref{fig:single_class_result}, our method adeptly models each flower type, faithfully capturing their distinct appearances.
This study underscores the potential of 3D-GPT in achieving precise object modeling and fine-grained attribute control of object types and visual characteristics.

\begin{figure*}[t]
    \vspace{-0.2cm}
    \centering
    \includegraphics[width=1\linewidth]{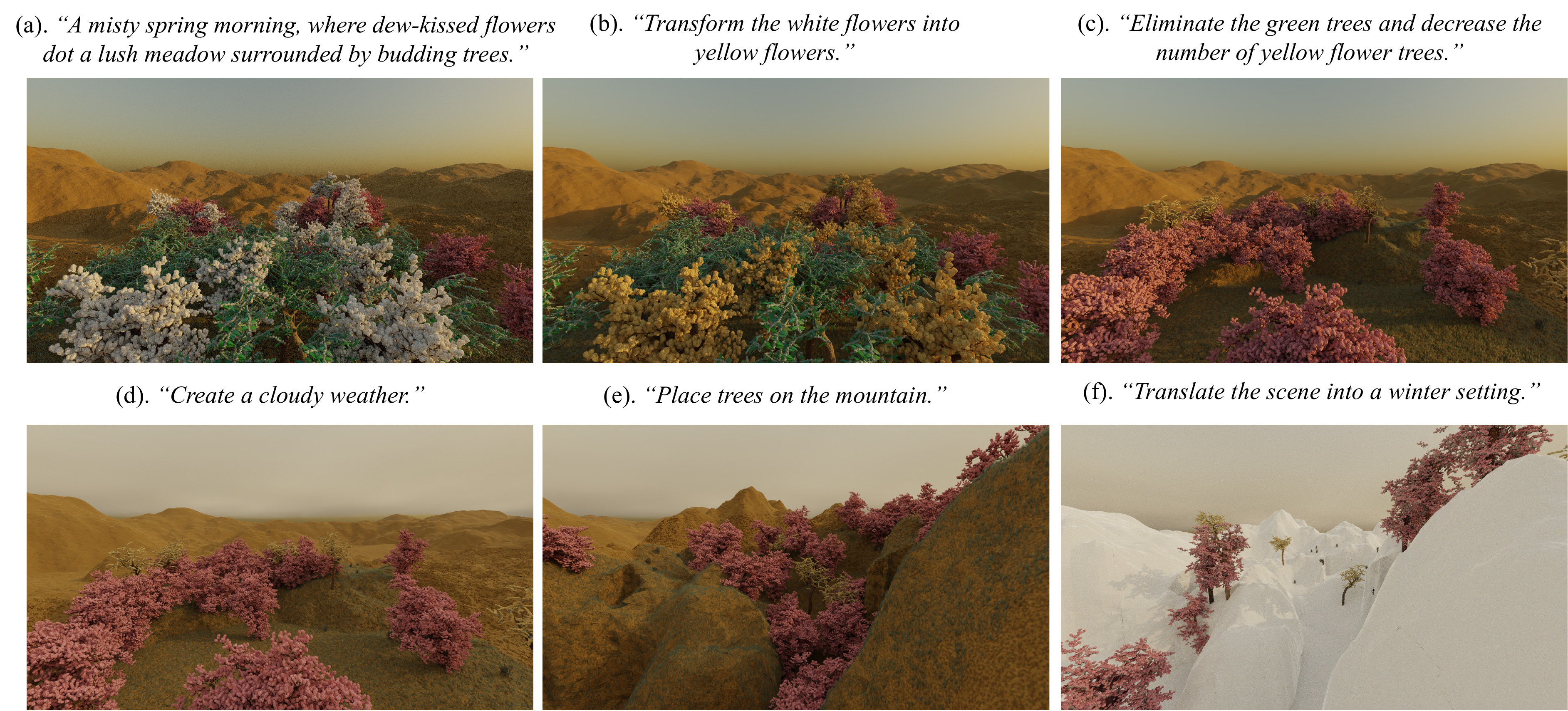}
    \caption{\textbf{Subsequence Instruction Editing Result.} (a) Initial instruction-generated scene. (b)-(f) Sequential editing steps with corresponding instructions. Our method enables controllable editing and effective user-agent communication.}
    \label{fig:scene_editing}
\end{figure*}

\paragraph{Subsequence Instruction Editing.} Here, we test the ability of 3D-GPT for effective human-agent communication and task manipulation. In Figure ~\ref{fig:scene_editing}, we observe that our method can comprehend subsequence instructions and make accurate decisions for scene modification.
Note that, unlike the existing text-to-3D methods, 3D-GPT maintains a memory of all prior modifications, thereby facilitating the connection of new instructions with the scene's context. Furthermore, our method eliminates the need for additional networks for controllable editings~\cite{zhang2023adding}.
This study underscores the efficiency and versatility of 3D-GPT in adeptly handling complex subsequence instructions for 3D modeling.

\paragraph{Individual Function Control.}

\begin{figure*}
    \vspace{-0.5cm}
    \centering
    \includegraphics[width=1\linewidth]{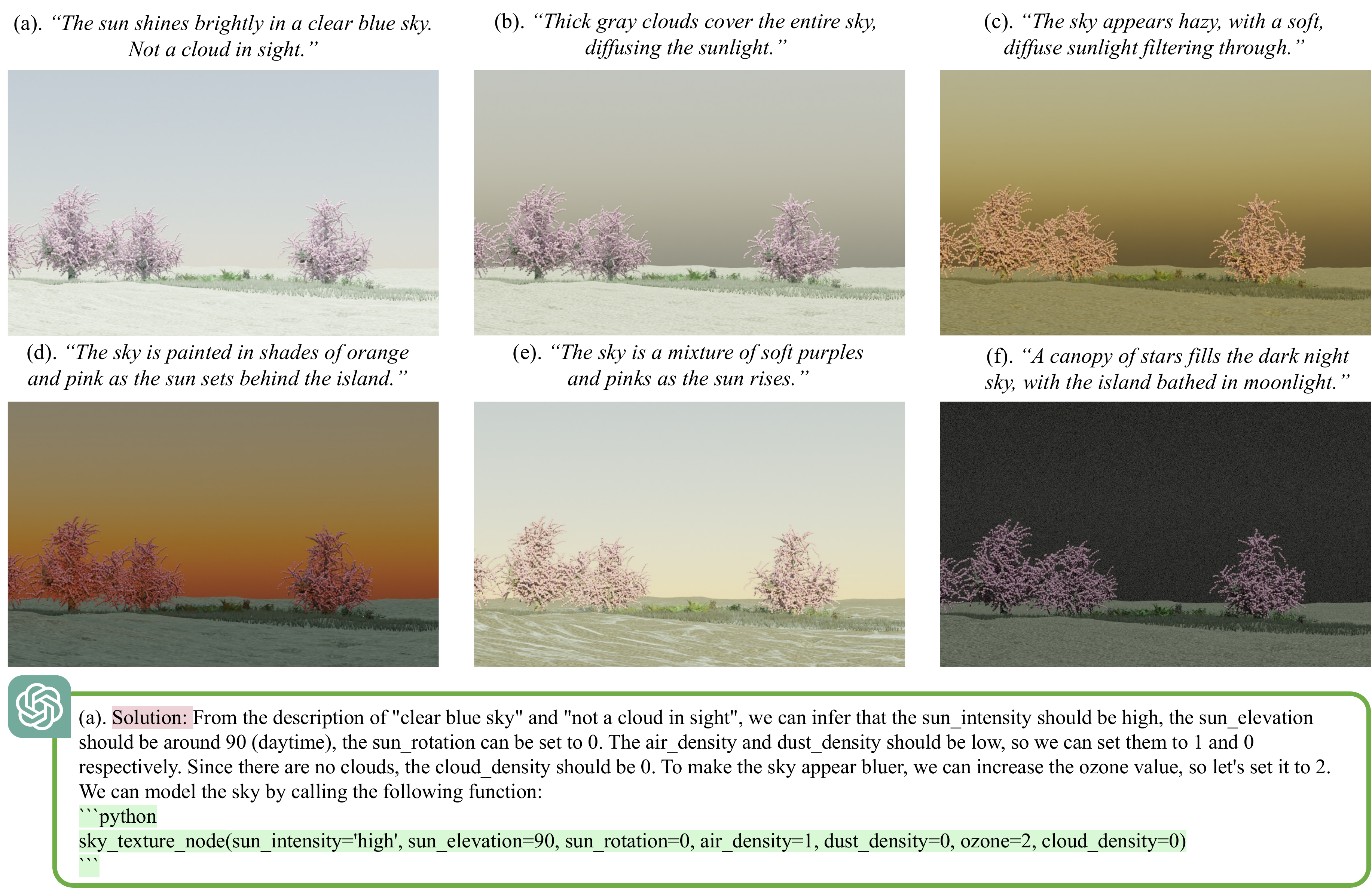}
    \caption{\textbf{Single Function Control Result.} Visual result (top) and modeling agent response example (bottom). Our method demonstrates a high degree of accuracy in inferring algorithm parameters, even when they do not possess a direct connection to visual appearance.}
    \label{fig:sky_modeling}
    \vspace{-0.5cm}
\end{figure*}

To evaluate the effectiveness of 3D-GPT in tool utilization, we present an illustrative example that highlights our method's ability to control individual functions and infer parameters.
Figure~\ref{fig:sky_modeling} exemplifies the capability of 3D-GPT to model sky appearances based on input text descriptions. It is worth noting that the function responsible for generating the sky texture does not directly correlate color information with sky appearance. Instead, it relies on the Nishita-sky modeling method, which requires a profound understanding of real-world sky and weather conditions, considering input parameters.
Our method adeptly extracts crucial information from the textual input and comprehends how each parameter influences the resulting sky appearance, as evident in Figure~\ref{fig:sky_modeling} (c) and (d). These results demonstrate that our method can effectively use individual functions as well as infer corresponding parameters.

\subsection{Ablation Study}

We conduct separate ablation studies for the Conceptualization Agent and Task Dispatch Agent. Our assessment focused on CLIP scores~\citep{radford2021learning}, failure rates, and parameter diversity, quantified using the categorical Shannon Diversity Index.
The CLIP score measures the alignment between text and generated images. The failure rate represents the percentage of system failures due to issues such as incorrect datatypes, wrong response patterns, or missing parameters from the Modeling Agent. Parameter diversity aims to gauge the diversity of generated outputs.

\begin{table}[h]
  \begin{subtable}{0.5\textwidth} 
    \centering
    \begin{tabular}{l|c}
      \toprule
      Metrics/ & CLIP    \\
      Method & Score \\
      \midrule
      w/o TDA &22.79\\
      Ours& 29.16\\
      \bottomrule
    \end{tabular}
    \caption{\textbf{Ablation Study of Task Dispatch Agent.}}
  \end{subtable}
  \begin{subtable}{0.5\textwidth} 
    \centering
    \begin{tabular}{l|ccc}
      \toprule
      Metrics/ & CLIP  & Failure & Parameter  \\
      Method & Score &  Rate & Diversity\\
      \midrule
    w/o CA &21.51&3.6$\%$&6.32\\
    Ours&30.30&0.8$\%$&7.34\\
      \bottomrule
    \end{tabular}
    \caption{\textbf{Ablation Study of Conceptualization Agent.}}
  \end{subtable}%
  \caption{\textbf{Ablation Study.}  ``w/o CA" indicates without the Conceptualization Agent,  ``w/o TDA" indicates without the Task Dispatch Agent.}
  \vspace{-0.5cm}
  \label{tab:ablation}
\end{table}

\paragraph{Case Study of Task Dispatch Agent.}

For the Task Dispatch Agent, the CLIP score is measured using $100$ initial scene descriptions, each appended with one additional subsequence instruction for each scene. Table~\ref{tab:ablation} (a) shows that without the Task Dispatch Agent, the CLIP score dropped from $29.16$ to $22.79$. It is important to note that the Task Dispatch Agent primarily impacts the performance of subsequence instructions, as all functions are utilized for the initial instruction. These findings underscore the pivotal role of the Task Dispatch Agent in managing communication flow.

\begin{figure*}
    \vspace{-0.7cm}
    \centering
    \includegraphics[width=1\linewidth]{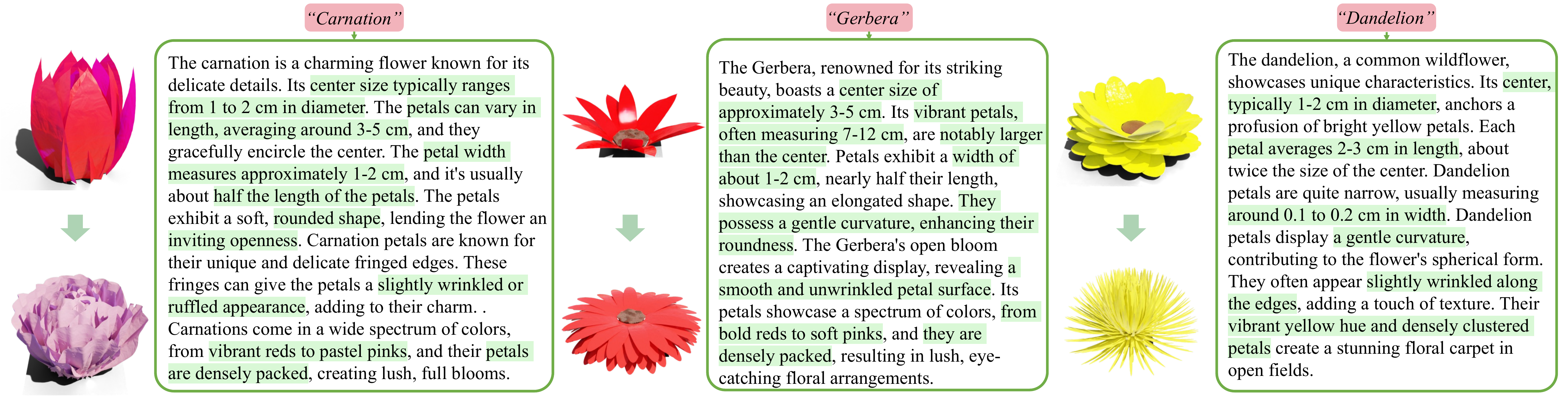}
    \caption{\textbf{Conceptualization Agent Case Study.} The enriched textual evidence demonstrates that the Conceptualization Agent provides essential knowledge for parameter inference (highlighted in green). For each subfigure, we compare the 3D model without (Top) and with (Bottom) agent. The models generated with the agent better match the text description than those without it.}
    \vspace{-0.40cm}
    \label{fig:case_study_1}
\end{figure*}

\paragraph{Case Study of conceptualization Agent.} For the Conceptualization Agent, the CLIP score is measured using $100$ initial scene descriptions. Table~\ref{tab:ablation} (b) displays the results, indicating that without the Conceptualization Agent, both text alignments (CLIP score) and parameter diversity decreased significantly. Moreover, the failure rate increased substantially, which adversely impacts the efficiency of the entire modeling process. Figure~\ref{fig:case_study_1} illustrates how the Conceptualization Agent facilitates the acquisition of essential knowledge for 3D modeling, providing a visual comparison of results with and without its involvement. When the Conceptualization Agent is engaged, the generated results closely align with the appearance of the intended flower type, highlighting its invaluable contribution to elevating overall 3D generation quality and fidelity.

\section{Discussion and Conclusion}

We have introduced 3D-GPT, a novel training-free framework for instruction-driven 3D modeling seamlessly integrated with procedural generation. Leveraging the capabilities of LLMs, 3DGPT aims to enhance human-AI communication in the context of 3D design. Our approach involves the collaborative efforts of three agents functioning as a cohesive 3D modeling team, ultimately yielding a 3D modeling file as output, as opposed to conventional 3D neural representations.
Moreover, our method consistently delivers high-quality results, showcases adaptability to expansive scenes, ensures 3D consistency, provides material modeling and editing capabilities, and facilitates real ray tracing for achieving lifelike visualizations.
Our empirical experiments show the potential of LLMs for reasoning, planning, and tool using in procedural 3D  modeling.
\paragraph{Limitations and Potential Directions.} While our framework has demonstrated promising 3D modeling results closely aligned with user instructions, it is essential to acknowledge several limitations:
1) Limited curve control and shading design: Currently, our framework lacks advanced capabilities for precise curve control and intricate shading design. Tasks involving the manipulation of tree branches or the blending of colors for leaf textures remain challenging.
2) Dependence on procedural generation algorithms: the effectiveness of our framework is contingent on the quality and availability of procedural generation algorithms. This reliance may limit results in specific categories, such as hair and fur.
3) Handling multi-modal instructions: challenges arise in processing multi-modal instructions, including audio and image inputs, potentially leading to information loss.
These limitations offer valuable insights for shaping future research and development in the field. We highlight three compelling directions for future investigation:
\\
\textbf{\textit{LLM 3D Fine-Tuning:}} It is promising to fine-tune LLMs to enhance their capabilities in geometry control, shading design, and fine-texture modeling. This refinement will make LLMs more adept at handling intricate 3D modeling tasks and grant greater creative control over the resulting 3D scenes.
\\
\textbf{\textit{Autonomous Rule Discovery:}} Building on the demonstrated tool-making capabilities of LLMs, one direction is to develop an autonomous 3D modeling system that reduces human involvement. This could empower LLMs to autonomously discover generation rules for new object classes and scenes, thus expanding the creative potential.
\\
\textbf{\textit{Multi-Modal Instruction Processing:}} To achieve more comprehensive and expressive 3D modeling based on varied user inputs, it is crucial to enhance the system's ability to comprehend and respond to multi-modal instructions. This would facilitate richer and more diverse 3D modeling outcomes, shaped by a broader spectrum of user inputs.
\\

\newpage
\bibliography{iclr2024_conference}
\bibliographystyle{iclr2024_conference}

\section{Appendix}

\subsection{Additional Result}

We kindly request the reader to consider visiting {\href{https://chuny1.github.io/3DGPT/3dgpt.html}{https://chuny1.github.io/3DGPT/3dgpt.html}} to view our high-quality 3D results.

\subsection{Prompt Example}

We offer an illustrative example of the prompt used for Adding Trees, presenting the Document, Code, Information, and a Usage Example.

\begin{figure*}[h]
    \centering
    \includegraphics[width=1\linewidth]{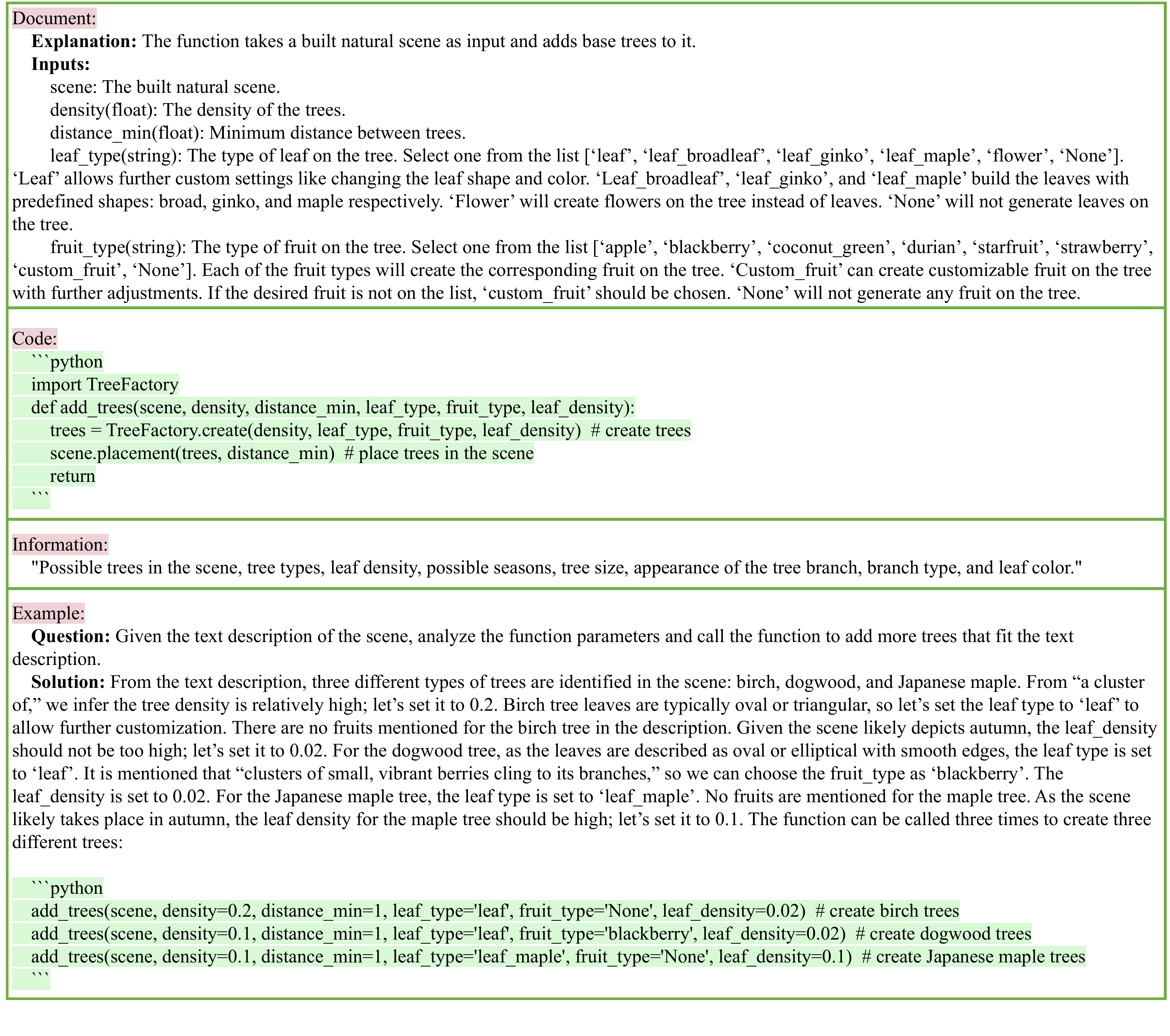}
    \caption{\textbf{Prompt Example of Adding Trees.}}
    \label{fig:prompt}
\end{figure*}

\begin{figure*}[h]
    \centering
    \includegraphics[width=1\linewidth]{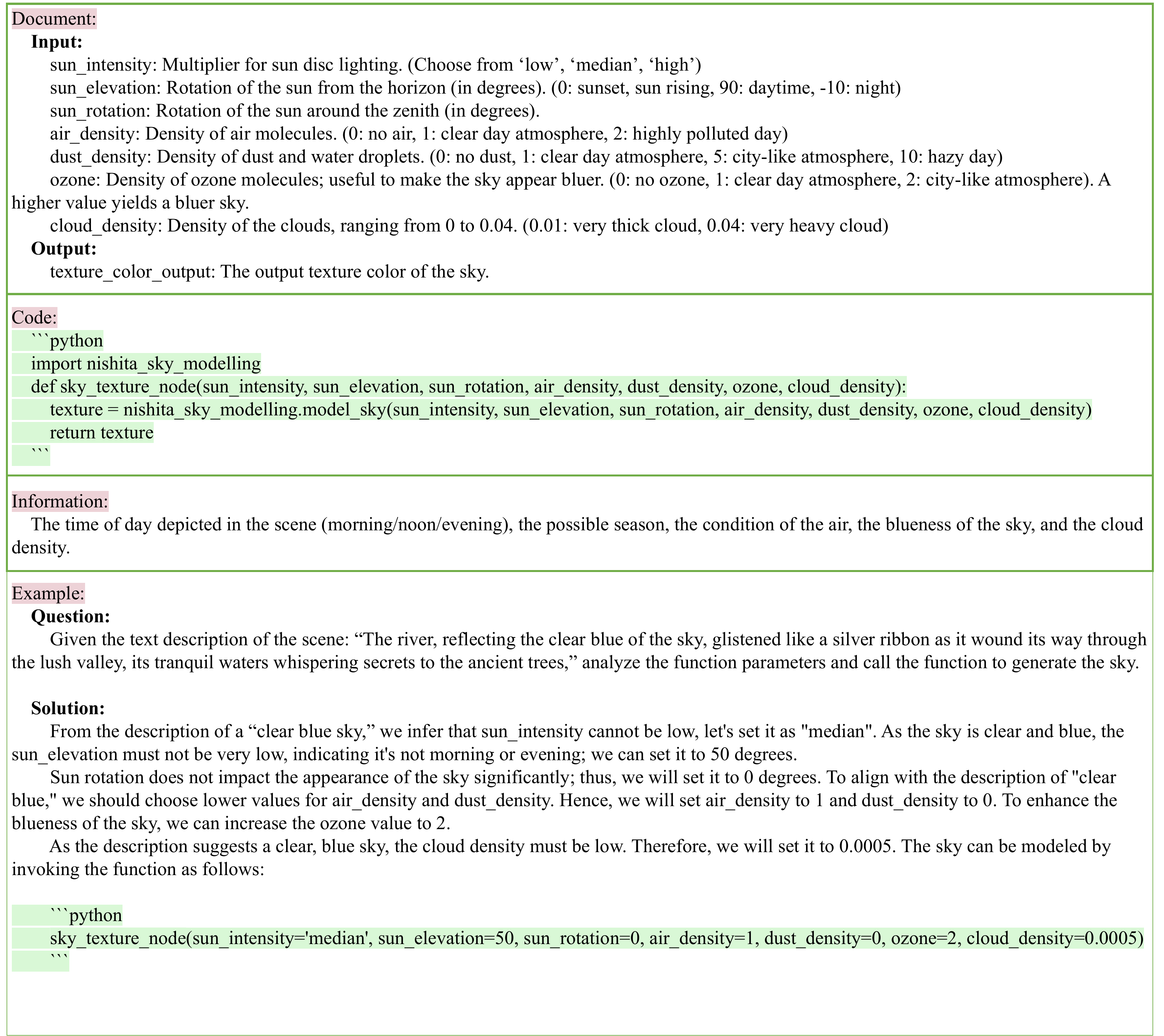}
    \caption{\textbf{Prompt Example of Sky Modeling Function.}}
    \label{fig:prompt2}
\end{figure*}

\subsection{Ablation Study Details}

We conduct separate ablation studies for the Conceptualization Agent and Task Dispatch Agent, evaluating their performance based on CLIP scores, failure rates, and parameter diversity. 

We use the implementation of~\cite{radford2021learning} to calculate CLIP score. It measures cosine similarity in the CLIP hidden space, indicating the alignment between generated images and text descriptions. For 3D scene CLIP scores, we placed a camera at the scene's center, capturing 50 images by rotating the camera 360 degrees. In the Conceptualization Agent evaluation, we used only the initial instruction. For the Task Dispatch Agent, both the initial instruction and one subsequence instruction were used.

The failure rate reflects the Modeling Agent's response. Failures can occur when the method cannot extract the correct pattern via the parser, generates data with an incorrect datatype, or omits/adds parameters to the function call. We evaluated five functions for each scene, recording the failure rate. We also assessed 500 parameters per scene, employing the Shannon Diversity Index to measure parameter diversity, and categorizing float parameters into 100 bins.

\subsection{Agent Responses and Visual Result}
\begin{figure*}[!ht]
    \centering
    \includegraphics[width=1\linewidth]{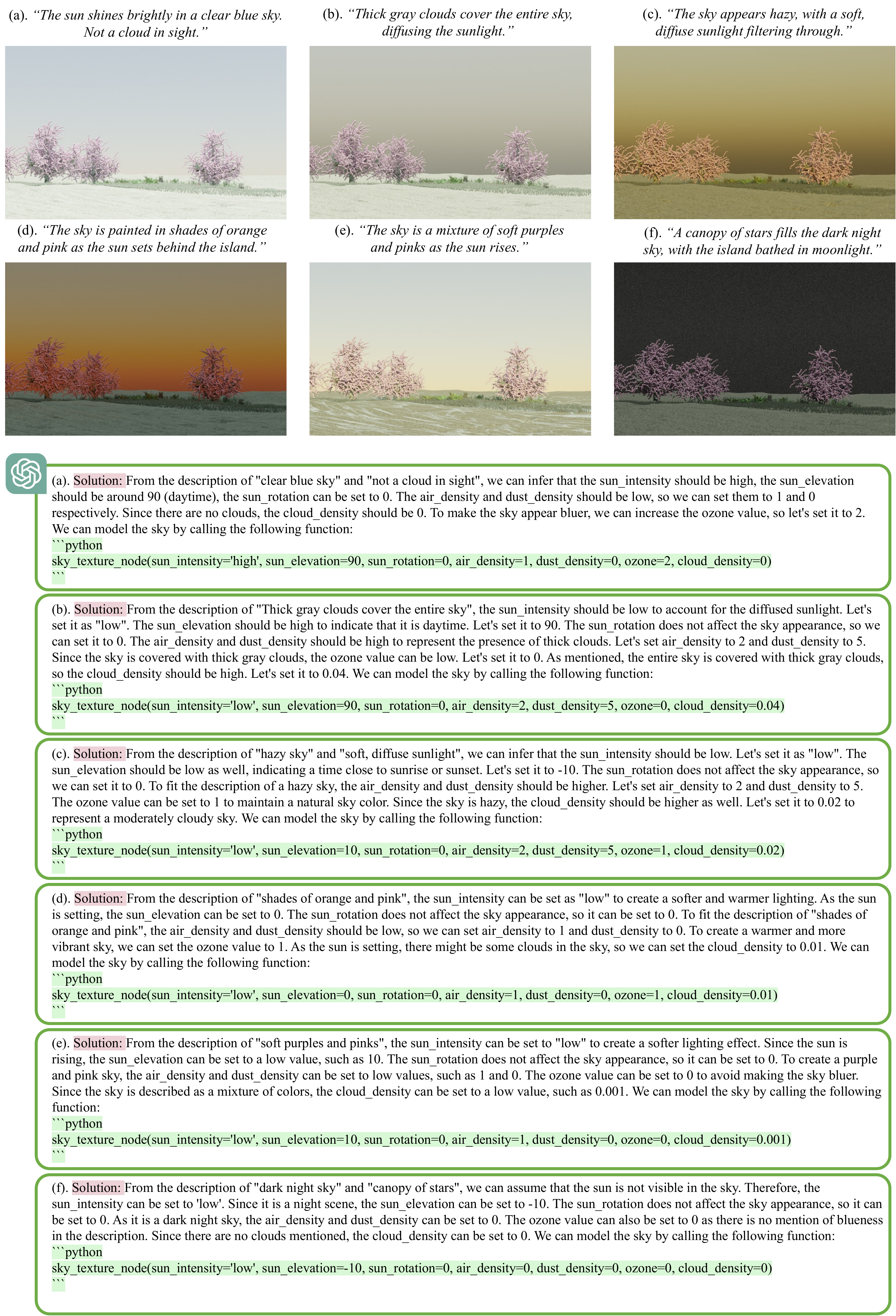}
    \caption{\textbf{Single Function Control Result.} Visual result (top) and modeling agent response example (bottom). Our method demonstrates a high degree of accuracy in inferring algorithm parameters, even when they do not possess a direct connection to visual appearance.}
    \label{fig:sky_modeling_full}
\end{figure*}

\end{document}